\pdfoutput=1
\documentclass[runningheads]{llncs}
\usepackage{graphicx}
\usepackage{multirow}

\begin{document}

\title{Robust Multimodal Image Registration using Deep Recurrent Reinforcement Learning \thanks{Supported in part by the National Natural Science Foundation of China under Grant 61602065, Sichuan province Key Technology Research and Development project under Grant 2017RZ0013, Scientific Research Foundation of the Education Department of Sichuan Province under Grant No.17ZA0062; J201608 supported by Chengdu University of Information and Technology (CUIT) Foundation for Leaders of Disciplines in Science, project KYTZ201610 supported by the Scientific Research Foundation of CUIT.}
} 
\titlerunning{Multimodal Image Registration using Reinforcement Learning} 


\author{Shanhui Sun\inst{1}\orcidID{0000-0001-9841-8592} \and
Jing Hu\inst{2}\orcidID{0000-0003-0921-0592} \and
Mingqing Yao\inst{2}\orcidID{0000-0002-5966-5440} \and
Jinrong Hu\inst{2}\orcidID{0000-0001-7732-8141} \and
Xiaodong Yang \inst{2}\orcidID{0000-0002-0973-2537} \and
Qi Song \inst{1} \and
Xi Wu \inst{2}\orcidID{0000-0002-7659-1631}
\thanks{The first two authors contributed equally to this paper}
}
%

\authorrunning{S. Sun et al.} 


\institute{CuraCloud corporation, U.S.A\and
Department of Computer Science, Chengdu University of Information Technology, P.R. China, 610225
}

\maketitle

\begin{abstract}
The crucial components of a conventional image registration method are the choice of the right feature representations and similarity measures. These two components, although elaborately designed, are somewhat handcrafted using human knowledge. To this end, these two components are tackled in an end-to-end manner via reinforcement learning in this work. Specifically, an artificial agent, which is composed of a combined policy and value network, is trained to adjust the moving image toward the right direction. We train this network using an asynchronous reinforcement learning algorithm, where a customized reward function is also leveraged to encourage robust image registration. This trained network is further incorporated with a lookahead inference to improve the registration capability. The advantage of this algorithm is fully demonstrated by our superior performance on clinical MR and CT image pairs to other state-of-the-art medical image registration methods.

\end{abstract}
\section{Introduction}
Image registration is a basic yet important pre-process in many applications such as remote sensing, computer-assisted surgery and medical image analysis and processing. In the context of brain registration, for instance, accurate alignment of the brain boundary and corresponding structures inside the brain such as hippocampus are crucial for monitoring brain cancer development \cite{author1}. Although extensive efforts have been made over three decades, image registration remains an open problem given the complexity clinical situations faced by the algorithms.

The core of image registration is to seek a spatial transformation that establishes pixel/voxel correspondence between a pair of fixed and moving images under rotation, scaling and translation transformation condition. Conventionally, the mapping between two images is obtained by minimizing an objective function with regard to some similarity criterion \cite{author1}. Therefore, two factors are fundamental: image feature representations and similarity measure \cite{author2}. Common image features include gradient, edge, geometric shape and contour, image skeleton, landmark, response of Gabor filter, or the intensity histogram. Recently, local invariant features have been widely applied to image registration \cite{author3}. As for similarity measure, sum-of-squared-differences, correlation coefficient, correlation ratio and mutual information are commonly used \cite{author4}. Three-dimensional extensions of these similarity measures have also been proposed to facilitate medical image registration.

Regarding the complexity of visual appearance of an image in presence of noise, outliers, bias field distortion, and spatially varying intensity distortion, defining the above-mentioned two factors appropriately is a challenging task. Although extensive studies have been carried out on the design of these two factors, they are somewhat designed manually and cannot be quite adaptable in the wide range of image modalities \cite{author1,author4,author5}. To this end, inspired by the success of neural architectures and deep learning for their strong semantic understanding, several works have generated image feature or similarity metric from scratch by using a deep convolutional neural network, without attempting to use any hand-crafted formations \cite{author6,author7,author8,author9,author10}.

\begin{figure}
\vspace{-0.2cm}  
\centering
\setlength{\belowcaptionskip}{-0.5 cm} 
\includegraphics[height=3.8cm]{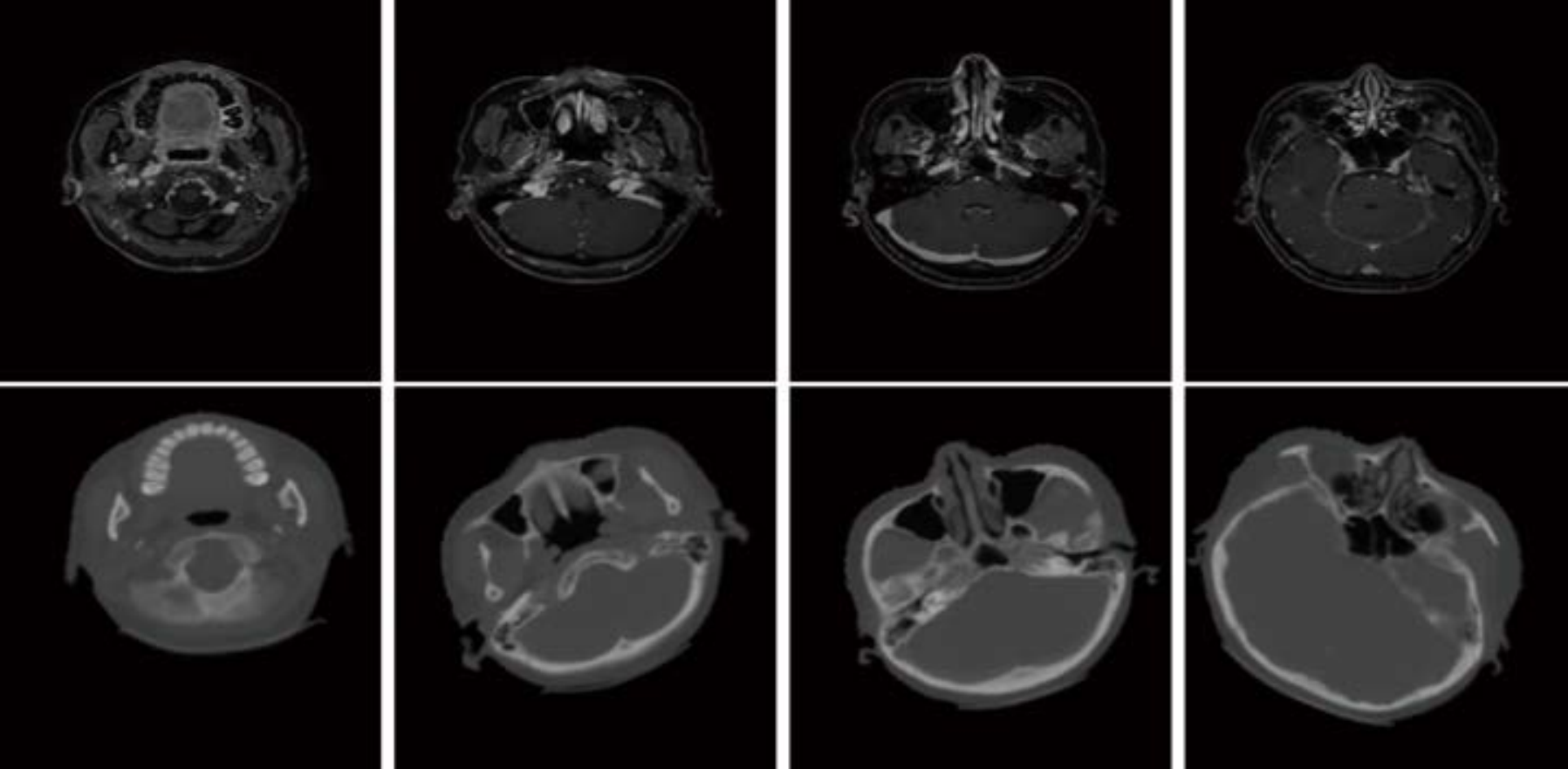}
\caption{Varying human anatomy in 2D multimodal image registration. Top: MR images (fixed image). Bottom: CT images (moving image). The goal of image registration is to align the moving image to the fixed image.}
\label{fig: varying anatomical}
\end{figure}

Recently, another type of approaches has also been proposed with focus on predicting registration parameters directly \cite{author5}. High parameter dimensionality and the non-linearity between image appearance and registration parameters render predictions non-trivial. More specifically, high dimensionality of data and parameter space (e.g. nine degree-of-freedom) challenge 3D medical image registration; while the challenges of 2D medical image registration are huge variability in appearances and shapes as indicated in Fig.~\ref{fig: varying anatomical}.

\begin{figure}[t]
\centering
\setlength{\belowcaptionskip}{-0.5cm} 
\includegraphics[height=5.5 cm]{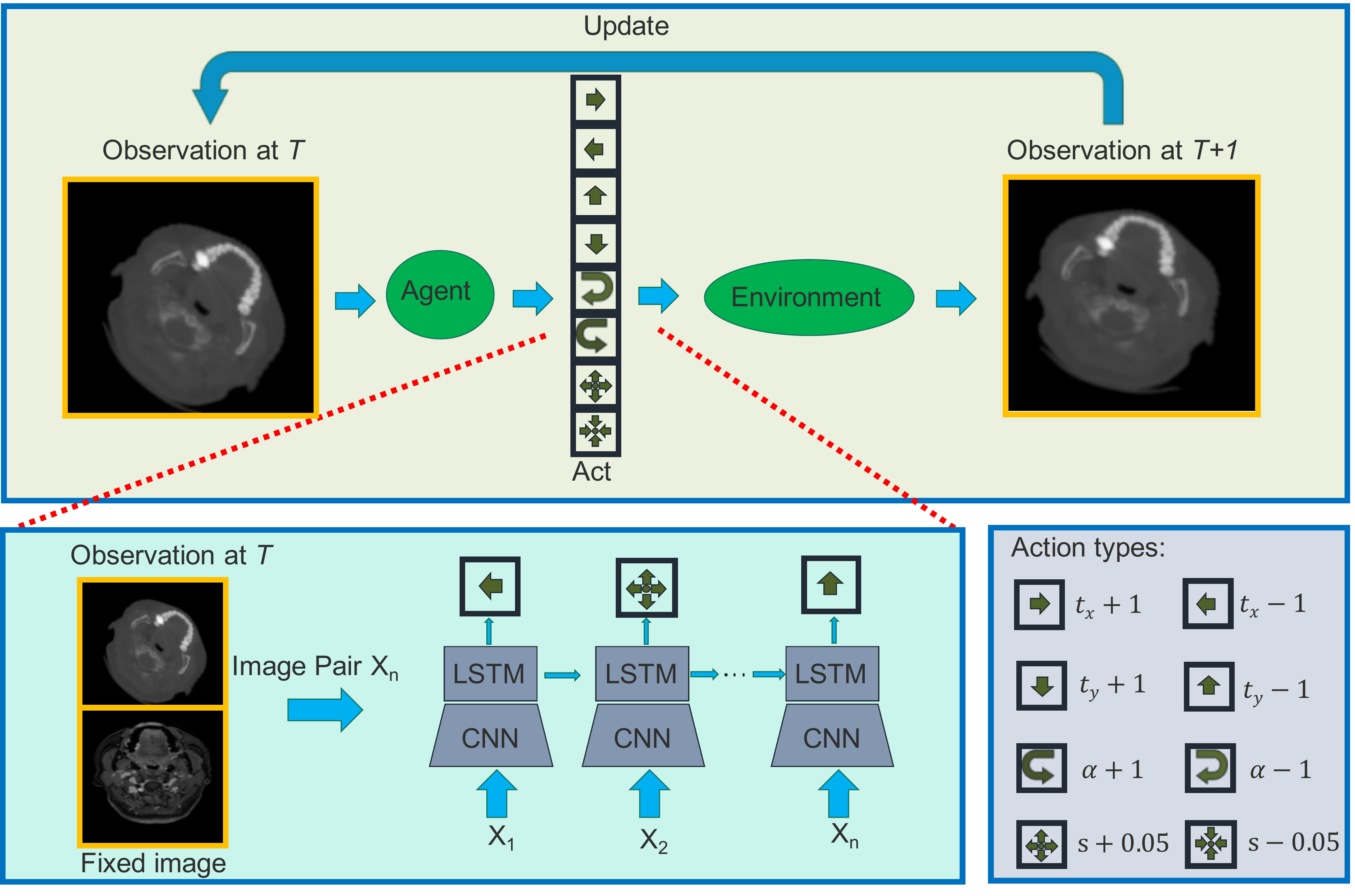}
\caption{Overview of our proposed framework. The upper part illustrates a sequential decision making driven by the agent who interacts with environment. The environment produces next observation image. The left bottom part depicts registration agent using CNN-LSTM network. The right bottom shows action types. }
\label{fig: framework overivew}
\end{figure}

Within the notion of the third type, we cast image registration as a sequential decision-making framework, in which both feature representations and similarity metric are implicitly learned from deep neural networks. An overview of our framework is shown in Fig.~\ref{fig: framework overivew}. An artificial agent explores the space of spatial transformation and determines the next best transformation parameter at each time step. Unlike other similar image registration methods that either use pre-trained neural network or learn the greedy exploration process in a supervised manner \cite{author11,author12}, our method explores the searching space freely by using a multithread actor-critic scheme (A3C). Our major contributions are:
\begin{itemize}
\item We present a novel reinforcement learning (RL) framework for image registration utilizing a combined policy and value network. This actor-critic network can explore transformation parameter spaces freely, thereby avoiding the local minima when a pose is far from initial displacement.
\item  To the best of our knowledge, this is the first RL-based image registration method that handle similarity transformation. To cope with the transformation parameter unit discrepancy, we introduce a new reward function driven by landmark error. Experiments also suggest that using such reward function helps convergence.
\item A Monte Carlo rollout strategy is performed as a look-ahead guidance in the test phase, whose “terminating” prediction is not trivial due to unknown terminal states.
\end{itemize}

\section{Related Work}
\subsection{Reinforcement learning}
In a classical reinforcement learning (RL) framework, an agent interacts with an environment in a sequence of observations, rewards and actions. At each time-step \(t\), the agent selects an action \(a_t\) from a set of pre-defined actions, so as to maximize the future rewards \(R_t\). The action is then passed to the environment and modifies its internal state and gives out immediate reward \(r_t\)  of this input action. This mechanism makes RL suitable for decision-making problems. The most impressive work is from DeepMind \cite{author13}, who has been able to train an agent to play the game of Go and achieve superhuman performance. Other research fields in computer vision also witness the success of RL \cite{author14,author15,author16,author17,author18}. For instance, an active object tracking system was proposed recently based on an actor-critic reinforcement learning model \cite{author16}, and this kind of actor-critic model was also validated to be effective in an image captioning scenario \cite{author17}.

In general, there are two types of RL methods: on-policy and off-policy. The former includes Sarsa, n-step method and actor-critic method and the latter includes Q-learning. In this paper, the framework of actor-critic is adopted. More specifically, the policy function is called an actor, which takes actions based on the current policy \(\pi \left( {{a_t}\left| {{s_t};\theta } \right.} \right)\); the value function is called a critic, which serves as a baseline to evaluate the quality of the action by returning the state value \(V\left( {{s_t};{\theta_v}} \right)\) for the current state under policy \(\pi \). To reduce the variance in policy gradient, an advantage function \(A\left( {{a_t};{s_t}} \right) = {R_t} - V\left( {{s_t};{\theta _v}} \right)\) is used for action \(a_t\) at state \(s_t\), where the expected future return \(R_t\) is calculated as a discounted sum of future rewards up to \(T\) time steps with discount factor \(\gamma  \in \left( {0,1} \right]\): \({R_t} = \sum\nolimits_{t' = t}^{t + T - 1} {{\gamma ^{t' - t}}{r_{t'}}} \). Parameters \(\theta \) of policy network and parameters \({\theta _v}\) of value network are updated as follows:
\begin{equation}
 \theta  \leftarrow \theta  + \tau {\nabla _\theta }\log \pi \left( {{a_t}\left| {{s_t}} \right.;\theta } \right)\left( {{R_t} - V\left( {{s_t};{\theta _v}} \right)} \right) + \beta {\nabla _\theta }H\left( {\pi \left( {{s_t};\theta } \right)} \right)
 \end{equation}
 \begin{equation}
    {\theta _v} \leftarrow {\theta _v} - \frac{1}{2}\tau {\nabla _{{\theta _v}}}{\left( {{R_t} - V\left( {{s_t};{\theta _v}} \right)} \right)^2}
  \end{equation}
where $\tau$ is the learning rate, $H\left(  \cdot  \right)$ is the entropy and $\beta$ is a regularization factor.

 Among the actor-critic frameworks, A3C \cite{author19} is the one that employs asynchronous parallel threads. Multiple threads run at the same time with unrelated copies of the environment, each generating its own sequences of training samples and maintaining an independent environment-agent interaction. The network parameters are shared across the threads and updated every \(T\) time steps asynchronously using Eqs. (1) and (2) in each thread. A3C is reported to be fast yet stable \cite{author16}.

\subsection{Image registration}
Early image registration processes can be roughly classified into two types: intensity-based methods and feature-based methods. The former, focusing on the image's gray spaces, maximizes the similarity between pixel intensities to determine the alignment between two images. Cross correlation and mutual information are considered to be gold standard similarity measures for this kind of methods. The latter extracts salient invariant image features and uses the correlation between those features to determine the optimal alignment. The main difficulty of early registration methods comes from the great variability of image when captured by different physical principles and environments, which translates in the lack of a general rule for images to be represented and compared. Motivated by the great advancement in computer vision triggered by deep learning technologies like convolutional neural networks (CNNs), several researchers have proposed to apply such techniques to the field of image registration. For example, Wu \textit{et~al.} \cite{author20} combined CNN with independent subspace analysis, the learned image features were then used to replace the handcrafted features in HAMMER registration model. Simonovsky \textit{et~al.} \cite{author7} employed CNNs to estimate a similarity cost between two patches from differing modalities.

 Recently, another way to formulate image registration problem is to directly predict transformation parameters \cite{author5}. Based on large deformation diffeomorphic metric mapping (LDDMM) registration framework, Yang \textit{et~al.} \cite{author21} designed a deep encoder-decoder network to initialize the momentum value for each pixel and then evolved over time to obtain the final transformation. Miao \textit{et~al.} \cite{author22} proposed a CNN-based regression approach or 2D/3D images, in which the CNNs are trained with artificial examples generated by manually adapting the transformation parameters for the input training data. However, this method requires a good initialization in proximity to the true poses. Liao \textit{et~al.} \cite{author11} formulated image registration as a sequential action learning problem and leveraged supervised learning to greedily choose alignment action at each exploration step for the sake of huge parameter searching space and the comparatively limited training data. However, the greedy searching may not be global optimal. Ma \textit{et~al.} \cite{author23}, the most relevant work to ours, extended work \cite{author11} via Q-learning framework trained using reinforcement learning. Although this work manages to search the transformation parameter space freely, a huge amount of state-action histories have to be saved during training. This becomes challenging when extended to 3D medical image registration.

\section{Method}

\subsection{Problem formulation}

Let \({{\bf{I}}_f}\) be a fixed image and \({{\mathbf{I}}_m}\) be a moving image. The task of image registration is to estimate the best spatial similarity transformation \({T_t}\) from \({{\mathbf{I}}_m}\) to \({{\bf{I}}_f}\). \({T_t}\) is parameterized by 4 parameters with 2 translations [\({t_x}\), \({t_y}\)], one rotation  \(\alpha \) and one scaling \(s\):
 \begin{equation}
 {T_t}\left( {{t_x},{t_y},s,\alpha } \right) = \left[ {\begin{array}{*{20}{c}}
  {s\cos \alpha }&{ - s\sin \alpha }&{{t_x}} \\
  {s\sin \alpha }&{s\cos \alpha }&{{t_y}}
\end{array}} \right]
\end{equation}

As illustrated in Fig.~\ref{fig: framework overivew}, we formulate this image registration problem as a sequential decision making process that at each time step, the agent decides which variable in \({T_t}\)  should be altered so that the moving image can be sequentially aligned to the fixed image. This process is modeled as a Markov Decision Process (MDP) with \(\left( {S,A,{r_t},\gamma } \right)\), where \(S\) is a set of states and \(A\) is a set of actions, \(r_t\) is the reward function the agent receives when taking a specific action at a specific state, \(\gamma \) is the discount factor that controls the importance of future rewards. In contrast to previous work \cite{author11} and \cite{author23}, we solved MDP in the framework of actor-critic using a new state approximation network, a novel reward function and a novel inference procedure. We focus on learning the policy function $\pi$ via deep reinforcement learning. More specifically, a new deep neural network is devised as a non-linear function approximator for $\pi$, where action $a_t$ at time $t$ can be drawn by $a_t\sim\pi(s_t;\theta)$, where $s_t$ is composed of ${\bf{I}}_m$ and ${\bf{I}}_f$ at time $t$.


\begin{table*}[t]
\begin{center}
\caption{Details of the combined policy and value network structure. C16-8-S4 represents 16 feature maps of convolution \(8 \times 8\) kernel with stride 4. FC256 indicates fully connected layer with 256 output dimensions. LSTM256 indicates that the dimension of all the elements (hidden state and input node) in the LSTM unit is 256.}
\label{table: paramters of network}
\begin{tabular}{l|c|c|c|c|c|c}
\hline
Layer &	1	& 2	& 3	& 4	& 5 & 	6\\
\hline
Parameter &	C16-8-S4	&C16-4-S2	&C32-4-MS-2	& FC256	& LSTM256 & Policy (8)/Value (1)\\
\hline
\end{tabular}
\end{center}
\end{table*}


\subsection{State Space Approximation and Action Definition}

\begin{figure}
\centering
\setlength{\belowcaptionskip}{-0.5cm}   
\includegraphics[height=5cm]{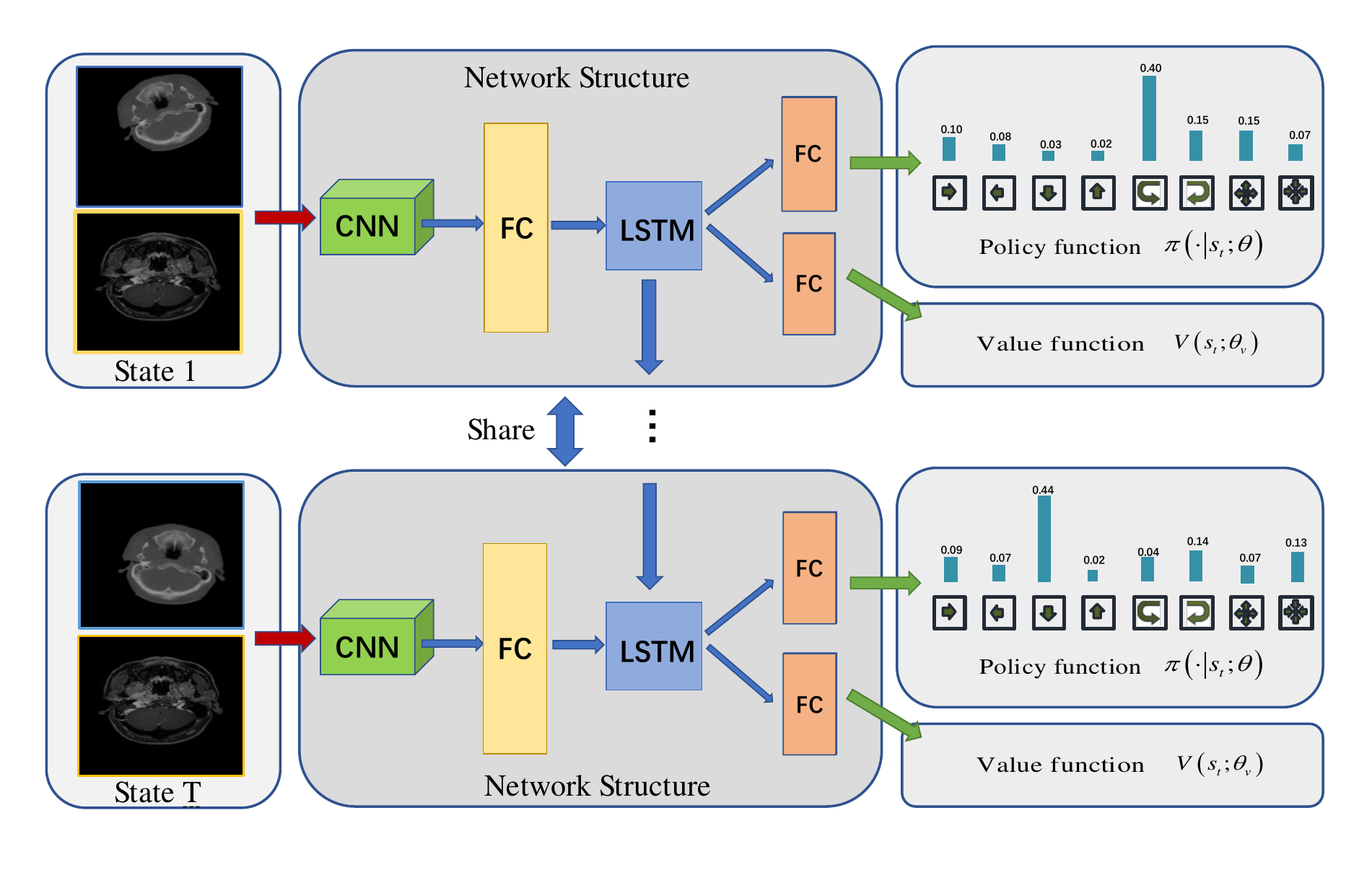}
\caption{The architecture of the proposed neural network. The network weights are shared for each LSTM unroll step and LSTM unroll $T$ times in our work.}
\label{fig: network architecture}
\end{figure}

In work \cite{author11} and \cite{author23}, state space is approximated using CNN and thereby neighbouring frames are not considered in the process of decision making. In our work, we utilize CNN to encode image states and a long short-term memory (LSTM) recurrent neural network to encode hidden states between neighbouring frames. Fig.~\ref{fig: network architecture} illustrates overview of the proposed network architecture. The input of the network is a concatenation of \({{\mathbf{I}}_m}\) and \({{\bf{I}}_f}\). The LSTM layer has two output (FC) layers: policy function, $\pi(s_t;\theta)$ and state value function, $v(s_t;\theta_v)$ . Table ~\ref{table: paramters of network} illustrates the details of our neural network. Each
convolutional layer as well as the fully connected
layer are followed by a eLU activation function. The action space consists of 8 candidate transformations that lead to the change of \( \pm 1\) pixel in translations, \( \pm {1^ \circ }\) for rotation, and \( \pm 0.05\) for scaling. At time \(t\), after the selection of action \(a_t\) by the agent, the transformation matrix \(T_{t+1}\) becomes \({a_t} \circ {T_t}\).

\subsection{Reward Definition}
Given a certain state, a reward is generated to reflect the value of current selected action. The definition of reward is somehow a tricky concept, as it mimics human learning. A wrong definition of reward will lead, with a high probability, to  unsatisfactory  learning results \cite{zhao2014full}. It is a natural intuition for most deep learning-based image registration methods that the reward function should encourage the transformation matrix, generated from the sequential actions made by the agent, to be closed to the ground truth. Therefore, the reward function in work \cite{author11} and \cite{author23} is inversely proportional to Euclidean distance between these two matrices.

However, considering scaling effect causes parameter unit discrepancy in our framework, since the scaling is in a comparatively small unit, and its parameter change is only \( \pm 0.05\) whereas the parameter change for translation and rotation is  \( \pm 1\). This implies that the Euclidean distance for scaling in two transformation matrices can be small, despite the rotation and translation being large. Therefore, a large reward is still obtained even scaling is not well registered. Although using a scalar to weight the contribution between scaling and other two parameters can mitigate this effect, the value of scalar should be carefully tuned. Instead of fine-tuning a specific weight function, we alternatively using landmark error to define the reward function. In our method, the key-points selected by scale invariant feature transform (SIFT) features \cite{author24} are viewed as landmarks. These landmark reference set \({{\mathbf{p}}_G}\) are computed from the ground truth of moving image. Afterwards, they are warped (\({{\mathbf{\tilde p}}_G}\)) using the perturbation transformation matrix. Then, for each action \(a_t\), the warped landmarks are transformed back using \(T_{t+1}\). The Euclidean distance \(D\) between the transformed landmarks and the corresponding ground truth is used to define the reward for action \(a_t\):
\begin{equation}
    {r_t} =  - D =  - \frac{1}{{{}^\# \left\{ {{{\mathbf{p}}_G}} \right\}}}\sum\limits_i {{{\left\| {{p_i} - {{\tilde p}_i} \circ {T_{t + 1}}} \right\|}_2}} ,{p_i} \in {{\mathbf{p}}_G},{\tilde p_i} \in {{\mathbf{\tilde p}}_G}
\end{equation}
where \({p_i}\) and \({\tilde p_i}\) are the landmark points, \( \circ \) denotes the align operator, \({}^\# \left\{ {} \right\}\) calculates the number of points. In addition, if \(D\) is smaller than a threshold, we assume that the terminal is triggered, and a terminal reward is set in this situation.
\subsection{Training using Reinforcement Learning}
We train the agent in an end-to-end manner using reinforcement learning algorithm similar to A3C proposed in \cite{author19}. However, rather than running different copies of a single game in \cite{author19}, each thread runs with a different image pair in our work. The episode starts from a pair of images and ends with a terminal state (image pair aligned) or reaches a maximum episode length ($M_e$) in our work. Since \(\pi \left( { \cdot \left| {{s_t};\theta } \right.} \right)\) and \(V\left( {{s_t};{\theta _v}} \right)\) are combined in our method, parameters of the agent are represented by \(\Theta  = \left\{ {\theta ,{\theta_v}} \right\}\). Gradients of \(\Theta\) are backpropagated from the actor-critic outputs to the lower-level layers. Note that different training episodes with similar initial poses would cause over-fitting, since small change of actions make two successive states strongly correlated. To handle this problem, we randomly perturb initial moving image at every training episode.

\subsection{Training using Supervised Learning}
For better comparison, we also train the neural network (to more specific, the policy network) via supervised learning. Under this strategy, the agent at each time step selects the “optimal” action \({a_t}\) that has the minimal distance for the new transformation \({a_t} \circ {T_t}\) to the ground truth transformation \({T_g}\) among all the legal actions. Recall that in reinforcement learning, the action \(a_t\)  is selected according to the output probability distribution over all legal actions. Therefore, in supervised learning strategy, only the shortest path for action space is explored.

To train the neural network, the policy network \(\pi \left( { \cdot \left| {{s_t};\theta } \right.} \right)\) is trained by minimizing cross entropy loss: \( - \sum\nolimits_{t = 1}^T {\log \pi \left( {{a_t}\left| {{s_t}} \right.;\theta } \right)} \) and the value network  \(V\left( {{s_t};{\theta_v}} \right)\) is trained by minimizing the mean squared loss \({\left( {{R_t} - V\left( {{s_t};{\theta_v}} \right)} \right)^2}\).
\subsection{Inference with Monte Carlo Rollout}
For training, the agent learns a registration policy that maps the current state to the optimal action. For inference in test, the agent applies the learned policy to approach the correct alignment. The stopping criteria in test phase is based on value network prediction $v_t\ge{trs}$, where $v_t$ is a predict value at time $t$ and $trs$ is a value threshold close to the terminal reward. However, it is not trivial to find a good $trs$ from the critic network as a termination since terminal state is unknown in the test phase, so this value can only be determined empirically on training data with new random perturbations. But, we often observe that either transformation parameters jitter around specific values or the terminal is not reachable in the test phase. To handle this challenge, a Monte Carlo (MC) method is proposed in the test phase to simulate multiple searching paths, so as to predict a better action. Note that we only perform MC rollout when $v_t$ reaches $trs$; otherwise the action is the one that with a highest policy probability.

To perform MC rollout, a number $N_{mc}$ of trajectories are simulated from the state \(s_t\), whose $v_t\ge{trs}$, with a fixed searching depth $D_{mc}$. For each trajectory \(i\), the agent at first randomly selects an action, then subsequent actions $a_{t'}$ are chosen following $a_{t'}\sim\pi(s_{t'};\theta)$ and at the same time obtained the corresponding values by value network. Suppose that the trajectory \(i\) finally gets a transformation matrix with parameters \(\left[ {{t_{xi}},{t_{yi}},{s_i},{\alpha_i}} \right]\). Moreover, accumulating the values along trajectory \(i\) obtains the total path value $V_{i}=\sum_{ri=t}^{t+D_{mc}}v_{ri}$, where $ri$ is an explored node at trajectory $i$. We computed the final parameters each based on weighted average:
\begin{equation}
  \left[ {{t_x},{t_y},s,\cos{\alpha},\sin{\alpha}} \right] = \frac{1}{{\sum\nolimits_{i=1}^{N_{mc}} {{V_i}} }}\sum\nolimits_{i=1}^{N_{mc}} {{V_i} \times \left[ {{t_{xi}},{t_{yi}},{s_i},\cos {\alpha_i},\sin {\alpha_i}} \right]},
\end{equation}
where $\alpha$ can be computed from $\cos{\alpha}$ and $\sin{\alpha}$.
The idea behind this formula is that it forms a Monte Carlo importance sampling along different trajectories, and the empirical mean results an approximation to the expected value.

\section{Experiment and Results}
\subsubsection{Dataset} In this paper, we apply our proposed framework to multi-modality image registration problem. The dataset used in our experiment contains 100 paired axis view MR images and CT images each from 99 patients diagnosed as nasopharyngeal carcinoma. The original CT and MR images have different resolutions, i.e. CT has the resolution of \(0.84 \times 0.84 \times 3\) mm and MR has the resolution of \(1 \times 1 \times 1\) mm. In term of image preprocessing, these CT images were resampled to an isotropic resolution of 1mm. In addition, since the ground truth alignment of these two data modalities is unfortunately not easily obtainable, the standard of alignment is estimated using an off-the-shelf toolbox Elastix \cite{author26} for the sake of efficiency.  More specifically, the MR image is treated as a fixed image while the CT image is treated as a moving image. Pre-registration for all MR and CT pairs for the same patient was carried out in 3D by Elastix with standard parameters. Taking into consideration that adjacent slices are strongly correlated, only six slices out of 100 are selected for each modality to exhibit considerable variation between images, which ends up with 594 MR-CT pairs for our experiments.

We randomly selected 79 patients as the training data resulting 474 image pairs, and 20 patients as testing data resulting 120 image pairs. In training stage, at each episode, the moving images are randomly perturbed with translation $[-25, 25]$ pixels with a step $1$ pixel for $x$ and $y$ axes respectively, a rotation $[-{30^\circ },{30^\circ}]$ with a step $1^\circ$, and a scaling factor $[0.75,1.25]$ with a step $0.05$. The training procedure thus sees $160000$ initial image pairs ($20000$ training episodes $\times{8}$ threads). For testing, two data sets were generated using different moving image perturbation distribution. Each moving image has 64 perturbations. As a result, we have 7680 image pairs in each data set. ``$E_1$" is a data set consisting of moving images that are perturbed using the same perturbation priors as those used for generating training data. ``$E_2$" is a data set consisting of moving images that are perturbed using a larger range: translation $[-30, 30]$ pixels for $x$ and $y$ axes respectively, a rotation $[-{45^\circ },{45^\circ}]$, and a scaling factor $[0.75,1.25]$. All the training images and testing images are resized to \(168 \times 168\) in our experiments.

\subsubsection{Methods used for comparison} To evaluate the effectiveness of our method, we compare it with a SIFT-based image registration method \cite{author24}, a pure SL method (named as "pure SL") which contains a feedforward network with CT and MRI as inputs and transformation matrix as output (used a same CNN network architecture as in our RL model), as well as several variations of the proposed framework so as to demonstrate our contributions in agent training, reward function and lookahead inference. Recall that in our method the policy and value networks are trained by RL (named as ``RL"), a landmark error based method is proposed for reward function (named as ``LME"), and Monte Carlo rollout strategy is used for prediction in the last step (named as ``MC"). Three corresponding substitutes are also used in later experiments, including agent trained with SL (named as ``SL"), reward function calculated by the transformation matrix distance (named as ``matrix"), and no lookahead inference. In general, the methods used in our experiments are: [SIFT, pure SL, RL-matrix, SL-matrix, RL-matrix-MC, SL-matrix-MC, RL-LME, SL-LME, RL-LME-MC (the proposed method), SL-LME-MC]. Note that in all experiments, target registration error (TRE) is used as a quantitative measure, which is the distance after registration between corresponding points not used in transformation matrix calculation \cite{author27}.

\subsubsection{Training and testing details} We trained the agent with 8 asynchronous threads, and with the Adam optimizer at an initial learning rate of 0.0001, \(\gamma  = 0.99\) and \(\beta  = 0.1\). Training episodes had the maximum length of 20000 cycles, and the maximum length ($M_e$) of each episode is 500 steps. Each episode is terminated whenever the agent reaches a distance terminal (in terms of landmark error or transformation matrix error) or the maximum episode length. The distance threshold for terminal is 1 for both kinds of errors, and the terminal reward in RL training is 10. The boostrapping length (the network weights updating frequency) in training procedure is 30. Each thread randomly selects a new training image pair at every two episodes.
Stopping thresholds $trs$ are RL-LME: 10, RL-LME-MC: 9, SL-LME: -0.05 and SL-LME-MC: -0.1.  We also applied these thresholds to their ``matrix" counterparts. For MC rollout methods, simulation number $N_{mc}=20$ and searching depth $D_{mc}=10$.


\subsubsection{Evaluation and Results} 
\begin{figure}
\centering
\setlength{\belowcaptionskip}{-0.5cm}
\includegraphics[height=7.5cm]{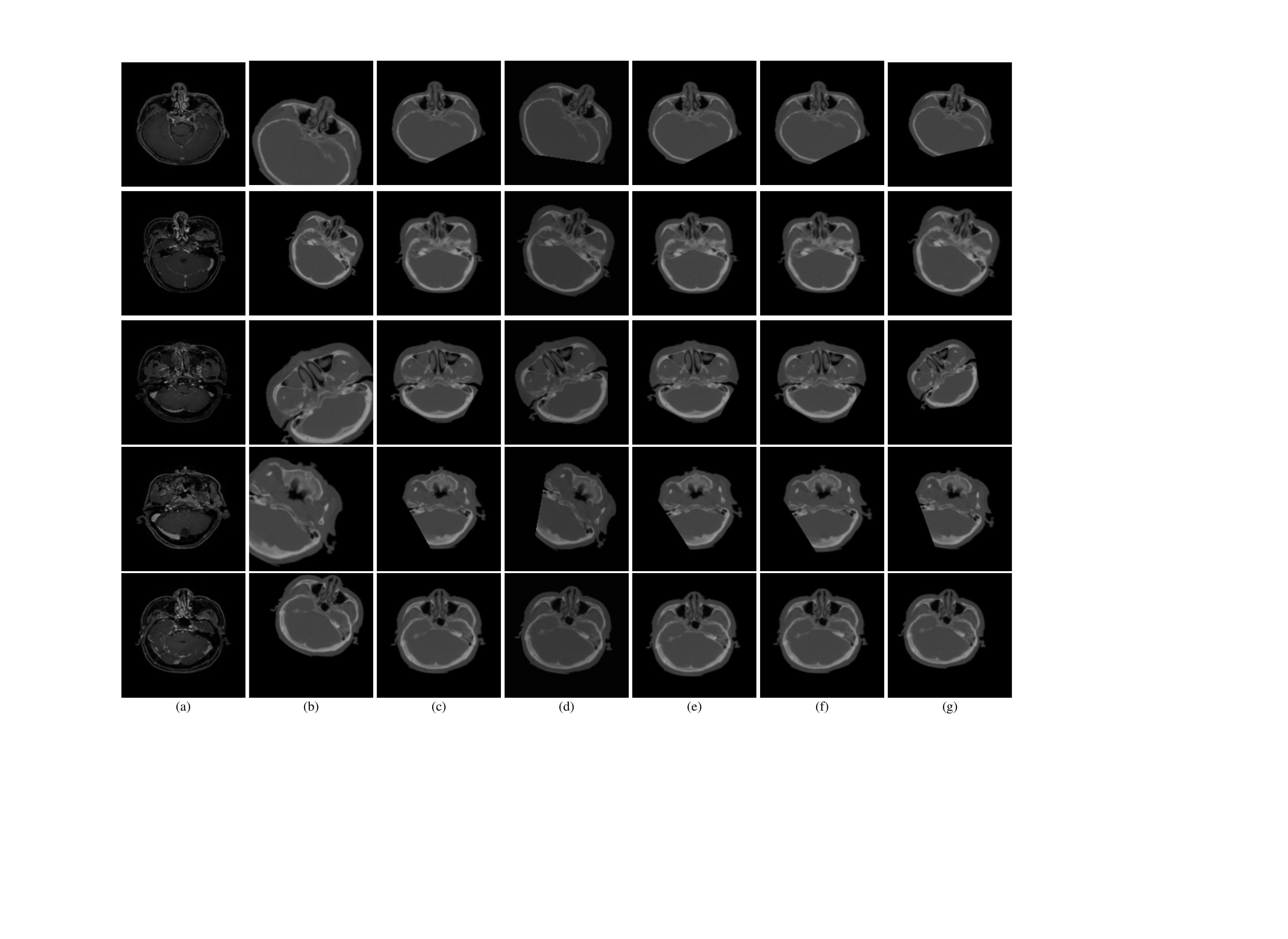}
\caption{Examples of registration results. (a) Fixed image, (b) initial moving image, (c) ground truth poses, results of (d)~Elastix (e)~SL-matrix, (f)~RL-LME-MC (Ours), (g) Pure ~SL.}
\label{figure:elastix comparison}
\end{figure}

We evaluated the models and inference variants (w/o MC) on datasets $E_1$ and $E_2$, respectively. Table~\ref{table:ComparisonBetweenMethods} summarizes the quantitative results. It is clear that all the deep-learning based image registration methods significantly outperform the method using SIFT. It is important to note that our method is far better than pure SL, which makes sense because the distance between the moving image and fixed image is so large that a direct regression learned by the latter is not feasible. The improved performance by SL-matrix, SL-matrix-MC and SL-LME suggests that using discrete action step by step is beneficial for the "bad" registration scenario.  Table~\ref{table:ComparisonBetweenMethods} also proves that these three components, including model “RL”, LME-based reward function and predication strategy, all contribute to the good performance of the proposed method. Furthermore, the proposed method is also compared with Elastix, which is used as a gold standard to pre-align the original MR-CT pairs for our experiments. Although we have found that this software can obtain a very accurate alignment for each image pair, the deformation between original MR-CT is in fact not severe. Is Elastix still a gold standard for our method when it applies to a more challenging data, i.e. the testing dataset used in our experiment, is worth looking into. Therefore, we performed a 2D version Elastix registration on $E_1$ and $E_2$. Note that 2D Elastix may yield registration results closer to ground truth, since the same cost function is used as the one used for ground truth generation. Such a comparison is made and summarized in Table ~\ref{table:ComparisonBetweenMethods}. A visual comparison is also presented in Fig.~\ref{figure:elastix comparison}. We can see that Elastix fails to register images with large deformation whereas our method is more robust to the challenging cases.

\begin{table}[t!p]
\center
\caption{Evaluation of methods on different datasets ($E_1$ and $E_2$) in terms of TRE. The statistics mean ($\mu$), standard deviation ($std$), median ($50^{th}$) and $90\%$ percentile ($90^{th}$) are performed. The bold numbers indicate the best performer in that column. Note that 2D Elastix results may exist bias due to it was performed using the same cost function as ground truth generation algorithm.}
\label{table:ComparisonBetweenMethods}
\begin{tabular}{l|lcc|lcc}
\multirow{2}{*}{} &\multicolumn{3}{c|}{$E_1$}& \multicolumn{3}{c}{$E_2$}\\ \cline{2-7}
&$\mu\pm{std}$ &$50^{th}$& $90^{th}$&$\mu\pm{std}$ & $50^{th}$& $90^{th}$
\\
\hline
SIFT & 11.79$\pm${9.81} & 9.04 & 19.99 & 12.15$\pm${9.89} &9.13 & 20.06
 \\
Elastix&\textbf{1.31}$\pm$\textbf{1.06} & \textbf{0.91} & \textbf{2.46} & 2.33$\pm${1.86} &\textbf{1.07} & 5.33
\\
Pure SL&4.59$\pm${3.00} & 3.72 & 8.28 & 6.62$\pm${5.11} &5.11 & 13.12
\\
SL-matrix &2.29$\pm${2.92}&1.81 &3.77 &2.99$\pm${4.72}&2.04 & 4.83
\\
SL-matrix-MC &2.88$\pm${4.52} & 2.12 & 4.39 &3.58$\pm${13.20} &2.20 & 5.17
\\
SL-LME &2.60$\pm${6.62} &1.34 & 3.15 & 3.06$\pm${7.12} &1.41 &4.35
\\
SL-LME-MC &3.10$\pm${41.67} &1.09&3.03 &3.16$\pm${20.69} &1.19&4.20
\\
RL-matrix &1.59$\pm${1.61}&1.32 &2.96 &1.95$\pm${3.55}&1.40 & 3.41
\\
RL-matrix-MC &1.31$\pm${1.64} &1.00 &2.38 &1.81$\pm${8.00} & \textbf{1.07} &3.21
\\
RL-LME &1.66$\pm${1.56} &1.30 &3.12 & 1.91$\pm${2.39} &1.41 &3.61
\\
RL-LME-MC (Ours) &1.39$\pm${1.45} &0.98 &2.59 & \textbf{1.66}$\pm$\textbf{2.30} & 1.08 & \textbf{3.17}
\end{tabular}
\end{table}

\section{Discussion}
From the above experiments, it is clear that the proposed RL-LME-MC image registration framework is more powerful and robust than the other methods, which validates the effectiveness of the advocated three components. But, besides these three factors, the network architecture (CNN+LSTM) is also worth looking into. In this section, the importance of LSTM is firstly discussed. After that, the above-mentioned three components of our proposed are respectively analyzed. Last, we also talk about the difference between our method and Elastix.
\subsection{Network architecture analysis}
Although there are several image registration methods that have been proposed so far, to the best of our knowledge, only one work that handles registration problem using RL solely. A Dueling Network framework was used in their work, focusing on learning the action value for the agent. Our method, however, alternatively uses A3C to learn both action probability and state value for the agent. In this way, there is no need for our method to use a replay-memory that hinders the future application of RL to 3D image registration in case of high dimensionality. But, lacking replay-memory requires our neural network should be elegant enough to obtain a good feature representation that facilitates neural network training.

 A CNN-LSTM network is proposed in this paper, where the LSTM is leveraged to learn the information (hidden state) among consecutive frames that are correlated in fact. A natural question is “how important is LSTM for our method”. To answer it, we trained a new neural network that substitutes LSTM with FC128 (named as “CNN-LME”). Fig.~\ref{fig: learning curve}(a) and (b) compares this new network with our proposed one in terms of agent's speed reaching terminals  and cumulative reward per episode. Clearly, incorporating LSTM into CNN facilitates a more efficient and stable convergence. We believe such efficiency attributes to the capability of LSTM unit that takes into account historical states when producing output. In addition, CNN-LME-MC results $2.61\pm{3.19}$ pixels on $E_1$ and $3.52\pm{5.70}$ pixels on $E_2$ in terms of TRE (mean$\pm$standard deviation). This result is worsen than our approach. 
 \subsection{Impacts of three components in our method}
 \subsubsection{Model RL} To assess the merits of RL, it is compared with a SL-based network using the same architecture. Although there are several reasonable doubts about the assert that RL surpasses SL, we have found in our study that using RL achieved higher registration accuracy than all the cases using SL, as reflected in Table  \ref{table:ComparisonBetweenMethods}. It should be noted that the advantage for RL is more obvious when handling $E_2$, the perturbation of which is larger than the training data. This implies that by exploring  more path enables RL a very strong learning ability and generalization ability.
\subsubsection{Reward function} 
\begin{figure}
\centering
\setlength{\belowcaptionskip}{-0.7cm}
\includegraphics[height=7cm]{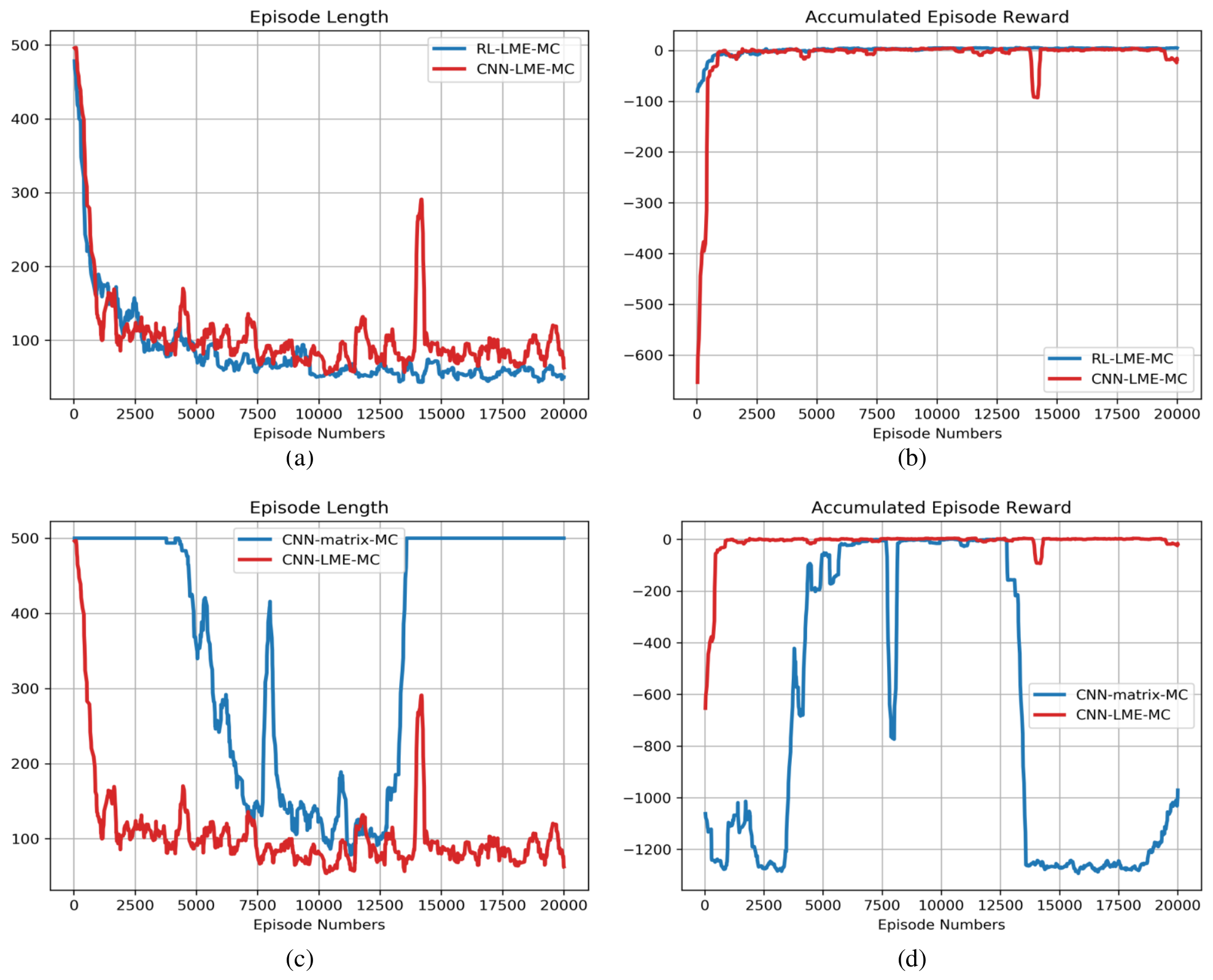}

\caption{Learning curves for different models. Steps per episode when agents reach terminals for (a) RL-LME vs CNN-LME and (c) CNN-matrix vs CNN-LME. Cumulative rewards per episode for (b) RL-LME vs CNN-LME and (d) CNN-matrix vs CNN-LME.}%
\label{fig: learning curve}%
\end{figure}

Reward function is a fundamental factor in RL. Although in Table \ref{table:ComparisonBetweenMethods}, no significant improvement of quantitative score is observed by using LME over matrix. It is even worse, in certain cases, that using LME obtained lower scores (see the third and fourth rows in Table \ref{table:ComparisonBetweenMethods} for details). However, we have found in our pilot study that for the model CNN-LME, if its reward function is changed into matrix, this new model (CNN-matrix) fails to converge and in fact very unstable within the same number of training steps, as shown in Fig.~\ref{fig: learning curve}(c) and (d). Although so far the mechanism behind this phenomenon is uncertain, it, on the hand, proves that defining a proper reward function can somewhat compensate for a weak network architecture.
\subsubsection{Lookahead} According to Table \ref{table:ComparisonBetweenMethods}, it is slightly better for a RL neural network than its SL counterpart to be embedded with a prediction step. This is possibly due to the reason that the action selected at each time step in RL is towards the goal of maximizing the future return. Incorporating a prediction step to RL would reinforce such intendancy. On the contrary, SL follows a greedy strategy that only considers the local optimal action. Hence, using a lookahead strategy implies averaging actions from future time steps, which would be highly likely to worsen the “optimal” situation compared with using the action at one time step. In addition, for ``MC", we made an early stopping (smaller $ths$) for compensating Monte Carlo computation expense. However, if the agent is stopped "too" early, using MC rollout may not reach an optimal solution. Due to this, an increased standard deviation can be observed at method RL-matrix-MC on $E_2$ data set. But, relaxing the threshold will address this problem.

\subsection{Comparison with Elastix}
Elastix \cite{author26} is a state-of-the-art medical image registration software that has been widely used in several research projects and it is still under active development. Both 2D and 3D image registration frameworks are provided, only dissimilar at image processing part. Based on the observation that Elastix performs well when the pose of moving image is in proximity to the fixed image's, its 3D version is used to generate the ground truth for our experiment. In comparison, the 2D Elastix could provide us a good reference to understand our method, although its results would be biased since it has a similar framework to the 3D algorithm. On data set $E_1$, 2D Elastix outweighs others. However, on data set $E_2$,  where we observed the pose of moving image is far from fixed image's, i.e. a large part of image missed, or in other words, a large portion is out of field of view, Elastix algorithm fails to fulfill the alignment task, as reflected in experiment results (Fig.~\ref{figure:elastix comparison} and Table~\ref{table:ComparisonBetweenMethods}). Quantitatively, Elastix performs much worse than our approach in terms of $90\%$ percentile on this large range testing data set. This is because Elastix is basically a least square optimization, and thereby is not robust to outliers and lack of high-level semantics understanding (e.g. parts, shapes and temporal context) which is crucial for robust alignment. In contrast, our proposed method performed similar on $E_1$ in terms of mean, standard deviation, median and $90\%$ percentile statistics of TRE, but much better on $E_2$. This implies that our method is able to perform semantic understanding as well as extrapolation although perturbation range is not included in the training prior.

\section{Conclusion}
In this paper, we present a new learning paradigm in the context of image registration based on reinforcement learning. Different from previous work that also build upon reinforcement learning, our method devised a combined policy network and value network to respectively generate the action probability and state value, without the need of extra storage on exploration history. To learn this network, we use a state-of-the-art reinforcement learning approach A3C with a novel landmark error based reward function. A Monte Carlo rollout prediction step is also proposed to further improvement. We evaluate our method with image slices from MR and CT scans, and find that our method achieves state-of-the-art performance on standard benchmark. Detailed analyses on our method are also conducted to understand its merits and properties. Our future work includes the sub-pixel image registration, extension to 3D registration, as well as the generalization to other computer vision problems.

%
%
%
%

\end{document}